\documentclass{ecai}
\usepackage{times}
\usepackage{graphicx}
\usepackage{latexsym}
\usepackage{amsmath,amssymb,amsfonts,nccmath,bm}
\usepackage{stackengine}
\usepackage{xspace}
\usepackage{caption}
\usepackage{subcaption}
\usepackage{comment}
\usepackage[hyphens]{url}
\usepackage{booktabs, multirow, multicol} 
\usepackage[psnames]{xcolor}
\usepackage[subtle]{savetrees}
\usepackage{hyperref}

\graphicspath{ {images/} }
\begin{document}

\title{$\mathtt{MedGraph:}$ Structural and Temporal Representation Learning of Electronic Medical Records}

\author{Bhagya Hettige\institute{Monash University, Australia, email: bhagya.hettige@monash.edu} \and 
        Weiqing Wang\institute{Monash University, Australia, email: teresa.wang@monash.edu} \and
        Yuan-Fang Li\institute{Monash University, Australia, email: yuanfang.li@monash.edu} \and 
        Suong Le\institute{Monash Health, Australia, email: suong.le@monashhealth.org} \and 
        Wray Buntine\institute{Monash University, Australia, email: wray.buntine@monash.edu}}


\maketitle
\bibliographystyle{ecai}

\begin{abstract}
  Electronic medical record (EMR) data contains historical sequences of visits of patients, and each visit contains rich information, such as patient demographics, hospital utilisation and medical codes, including diagnosis, procedure and medication codes. 
  Most existing EMR embedding methods capture visit-code associations by constructing input visit representations as binary vectors with a static vocabulary of medical codes.
  With this limited representation, they fail in encapsulating rich attribute information of visits (demographics and utilisation information) and/or codes (e.g., medical code descriptions). 
  Furthermore, current work considers visits of the same patient as discrete-time events and ignores time gaps between them.
  However, the time gaps between visits depict dynamics of the patient's medical history inducing varying influences on future visits. 
  To address these limitations, we present $\mathtt{MedGraph}$, a supervised EMR embedding method that captures two types of information: (1) the visit-code associations in an attributed bipartite graph, and (2) the temporal sequencing of visits through a point process. 
  $\mathtt{MedGraph}$ produces Gaussian embeddings for visits and codes to model the uncertainty. 
  We evaluate the performance of $\mathtt{MedGraph}$ through an extensive experimental study and show that $\mathtt{MedGraph}$ outperforms state-of-the-art EMR embedding methods in several medical risk prediction tasks.
\end{abstract}


\section{Introduction}

Electronic medical records (EMR) contain rich clinical data from a patient's stays in hospital. 
A high volume of EMRs is collected by hospitals that can be used in medical risk prediction to improve the quality of personalised healthcare.
EMR data forms a unique and complex data structure.
An EMR represents a hospital visit, and it typically contains patient demographics (e.g.\ age, gender) and hospital utilisation information (e.g.\ duration/ward of stay). 
Also, an unordered set of medical concepts (e.g.\ diagnosis, procedure and medication codes) are associated with each visit.
These medical concepts are usually taken from pre-defined standards in healthcare such as International Classification of Diseases (ICD) and National Drug Codes (NDC).
Moreover, EMRs accumulated over a period of time naturally form a sequence of visits of a patient's hospitalisation history.
Transforming EMRs into low-dimensional vectors has been an active research topic recently, as it enables these complex data in downstream machine learning algorithms to perform predictive healthcare tasks~\cite{DBLP:conf/ijcai/med2,DBLP:conf/nips/mime,DBLP:conf/kdd/dipole,miotto2016deep_pat,nguyen2016mathtt_deepr,DBLP:conf/ijcai/med1}.
Learning these \emph{structural} visit-code associations and \emph{temporal} visit-sequence influences are two important aspects of EMR embedding.

Considering structural visit-code associations,
a patient's visit contains a set of unordered medical codes.
Existing methods, such as Med2Vec~\cite{DBLP:conf/kdd/med2vec}, RETAIN~\cite{DBLP:conf/nips/retain} and Dipole~\cite{DBLP:conf/kdd/dipole}, propose sophisticated deep learning models to derive latent representations for visits using multi-hot-encoded medical codes as inputs. 
The major limitations of these methods are three-fold.
First, the visits are represented using a fixed vocabulary of medical codes, but in the real-world EMR systems, new or previously unseen medical codes can be introduced (due to revised versions of medical codes, e.g.\ ICD-9 and ICD-10).
Second, some medical codes are rarely reported and the existing methods does not deal with the sparsity of the data.
Third, they do not capture demographics and utilisation information attached with visits (except for Med2Vec~\cite{DBLP:conf/kdd/med2vec}), and side information found in medical codes such as ICD text code descriptions.
These additional attribute information are important in determining similar visits and similar medical codes.

Considering temporal visit sequences,
EMRs are longitudinal medical events which are time-stamped. 
Each patient has a temporal sequence of visits, and these visits possess time-dependent relationships among them, i.e.\ previous visits in a patient's history can have an influence on the next visit. 
The influence of historical visits on the next visit also degrades with time, so that more recent visits may have higher influence than older visits. 
Moreover, the time gap between consecutive visits is not fixed and the larger the time gap between two visits, the less related they are. 
Recurrent neural network (RNN)-based architectures~\cite{DBLP:conf/nips/retain,DBLP:conf/nips/mime,DBLP:conf/kdd/dipole,DBLP:conf/pakdd/deepcare} have been proposed in previous work to learn the temporality of visits, but they consider the visit sequences as time-series by treating the time as indexes (i.e.\ events are ordered periodically) and do not account for the varying time gaps.
On the contrary, visits are continuous-time events and there are varying influences of historical visits on the current visit, based on the time gaps between them.

\begin{figure}[!t]
    \centering
    \includegraphics[width=\linewidth]{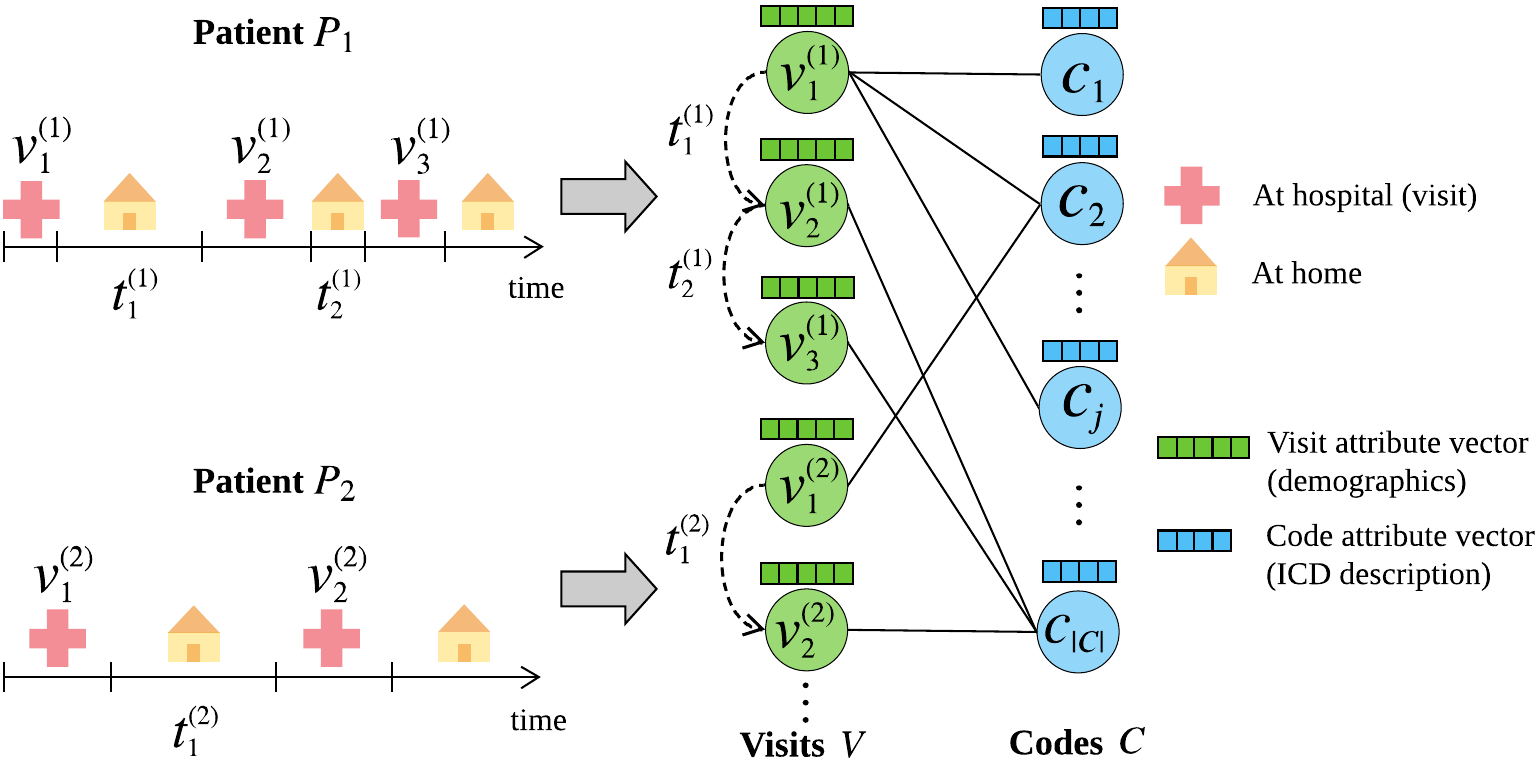}
    \caption{The $\mathtt{MedGraph}$ data structure. $v^{(i)}_j$ is the $j$-th visit of $i$-th patient.}
    \label{fig:graph}
    \vspace{-2mm}
\end{figure}

In this paper, we propose $\mathtt{MedGraph}$, a novel EMR embedding method that leverages both \emph{structural} and \emph{temporal} information in EMRs to improve the embedding quality. 
$\mathtt{MedGraph}$ encompasses a novel graph-based data structure to represent EMR, in which we represent visits and codes as nodes, and their different interactions as edges. 
In Figure \ref{fig:graph}, each patient has a temporal visit sequence connected via dashed, directed edges, and each visit has a set of codes connected via thick, undirected edges.
Denoting the sets of visits and codes as $V$ and $C$ respectively, $V$-$C$ relationships form a bipartite graph with $V$ and $C$ node partitions. 
Each node carries supplementary information such as demographics and utilisation information in visits, and textual descriptions in codes, which makes $V$-$C$ an attributed bipartite graph.
The $V \xrightarrow{t} V$ relationships form temporal sequences of nodes where $t$ is the time gap between two consecutive visits of a patient. 
Since the graph is an open data structure, it is extensible for new or unseen medical codes.

Taking advantage of the graph structure, $\mathtt{MedGraph}$ effectively learns low-dimensional representations for visits and codes by capturing both visit-code associations (based on the graph structure) and hospitalisation history (based on the temporal visit sequences) under one unified framework.
Structurally, $\mathtt{MedGraph}$ learns representations for visits and codes by considering them in an attributed bipartite graph. 
$\mathtt{MedGraph}$ proposes to model temporal visit sequence information as a temporal point process, such as the Hawkes process~\cite{hawkes_process}, 
to effectively capture and account for the varying influence of historical visits of different time gaps. 
These theoretical point process models often make strong assumptions about the generative process of the sequential event data which do not necessarily depict the real-world dynamics and also limit the expressive power of the model~\cite{DBLP:conf/kdd/rmtpp,point_processes}.
To automatically learn a more expressive representation for the varying historical visit influence when a real parametric model is unknown, $\mathtt{MedGraph}$ uses an RNN-based architecture, as in RMTPP~\cite{DBLP:conf/kdd/rmtpp}, to model the conditional intensity function.
We have designed $\mathtt{MedGraph}$ to be accountable for the uncertainty of the embeddings by learning node embeddings (i.e.\ visits and codes) as probability distributions (e.g.\ Gaussians). 
To the best of our knowledge, $\mathtt{MedGraph}$ is the first EMR embedding method that models uncertainty of the visit and code embeddings as Gaussian distributions.
The contributions of our proposed EMR embedding framework, $\mathtt{MedGraph}$, are threefold:
\begin{enumerate}
    \item A customised graph-based data structure designed for EMR data, that naturally captures both the visit-code co-location information ($V$-$C$) \emph{structurally} as an attributed bipartite graph and the visit sequence information ($V \xrightarrow[]{t} V$) \emph{temporally} as a continuous event sequence. 
    \item A novel approach to effectively and efficiently learn $V \xrightarrow{t} V$ relationships using a point process based sequence learning method with RNN-based conditional intensity function, considering varying influence of historical visits on the next visit.
    \item An extensive experimental study on two real-world EMR datasets that shows $\mathtt{MedGraph}$'s superiority over state-of-the-art embedding methods on a number of tasks: 30-day readmission prediction, mortality prediction, medical code classification, medical code visualisation and uncertainty modelling.
\end{enumerate}


\section{Related Work}

Representation learning of EMRs aims at learning low-dimensional vectors for hospital visits and medical codes through their interactions in terms of visit-code relationship and visit-visit sequencing information. 

Several embedding models have been proposed to learn from visit-code relationships in EMR data~\cite{DBLP:journals/titb/deep_ehr}.
Most of these works, including Med2Vec~\cite{DBLP:conf/kdd/med2vec}, RETAIN~\cite{DBLP:conf/nips/retain}, Deepr~\cite{nguyen2016mathtt_deepr},
GRAM~\cite{DBLP:conf/kdd/gram} and Dipole~\cite{DBLP:conf/kdd/dipole}, capture the visit-code associations by constructing the input visit vector as a multi-hot encoded medical code vector.
MiME~\cite{DBLP:conf/nips/mime} assumes the hierarchical structure of EMR data and represents patients, visits, diagnoses, procedures and medications in a hierarchy in the stated order.
In the real-world EMRs, though, this hierarchical granularity of medical codes is often not found. 
GCT~\cite{conf/aaai/GCT} attempts to address this challenge with a graph data structure to learn the implicit hierarchical relations of codes using Transformers.
However, none of these works consider the rich attributes of visits/codes. 
An exception to this are Med2Vec~\cite{DBLP:conf/kdd/med2vec}, which considers visit demographics, and MNN~\cite{DBLP:conf/ijcai/temp1}, which incorporates clinical notes. 
Still, they do not consider code attributes. 
In contrast, GRAM~\cite{DBLP:conf/kdd/gram} models EMR with a convex combination of the embeddings of the code and its ancestors on the ontology tree, and it strictly depends on this ontology structure.
But not all the medical codes form such rich ontology graphs.

Existing EMR representation learning work captures the temporality of visit sequences using either Skip-gram~\cite{DBLP:conf/nips/word2vec} with multi-layer perceptron (MLP)-based architectures~\cite{DBLP:conf/ijcai/time_aware_emr,DBLP:conf/kdd/med2vec} or RNN-based architectures~\cite{DBLP:conf/mlhc/doctorai,DBLP:conf/nips/retain,DBLP:journals/jamia/rnn_emr,DBLP:conf/nips/mime,DBLP:conf/naacl/rnn_med,DBLP:conf/kdd/dipole,DBLP:conf/pakdd/deepcare,DBLP:conf/ijcai/temp2}. 
However, Skip-gram is only capable of capturing neighbouring visits within a predefined number of time steps without any particular order in the context window. 
RNN models assume that the events are recorded periodically, and cannot effectively capture the varying time gaps between visits.
A recent work, PacRNN~\cite{DBLP:conf/ijcai/temp1}, proposes a continuous-time model using a point process for a specialised healthcare task, i.e.\ ranked diagnosis code recommendation and time prediction of the next visit. 

$\mathtt{MedGraph}$ uses a graph-based data structure to model EMR data, and we show that the graph can capture much 
more useful and meaningful data compared to the existing approaches.
Our approach also effectively captures the temporality of historical visits of the patients via the proposed temporal point process model.
Experimental results show that $\mathtt{MedGraph}$ produces more effective EMR embeddings.


\section{\textbf{$\mathtt{MedGraph}$}: Medical Data Graph Embedding}
\label{sec:method}

In this section, we describe the algorithm for $\mathtt{MedGraph}$.
Without loss of generality, the algorithm will be discussed for a single patient for simplicity of the notations.
Figure~\ref{fig:graph} is a heterogeneous graph with two types of nodes and two types of edges.
These two edge types denote two distinguishing types of information about the visits, i.e.\ codes in the visits and temporal visit sequences.
Therefore, we dismantle the two edge types.
Accordingly, we extract a subgraph from Fig.~\ref{fig:graph} for a single patient, and decompose it into an attributed bipartite graph (Fig.~\ref{fig:a}) and a temporal (timestamped) sequence graph (Fig.~\ref{fig:b}).

\subsection{Notations of $\mathtt{MedGraph}$}

\begin{figure*}[t]
  \centering
  \def\figa{\includegraphics[width=1.2in]{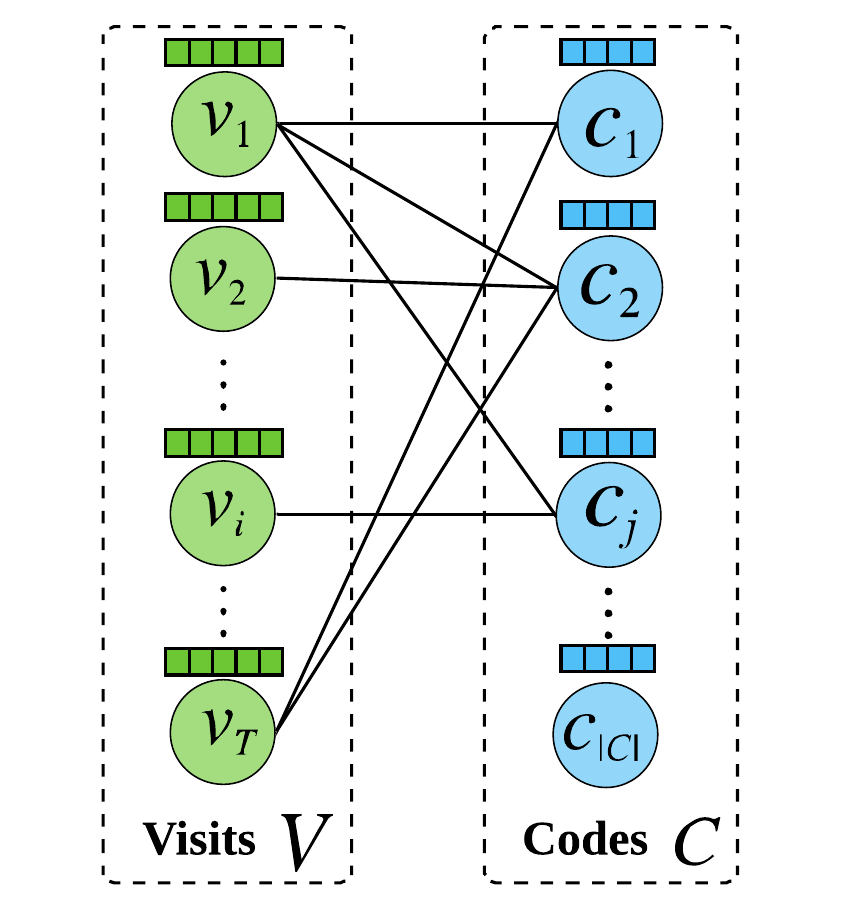}}
  \def\figb{\includegraphics[width=1.2in]{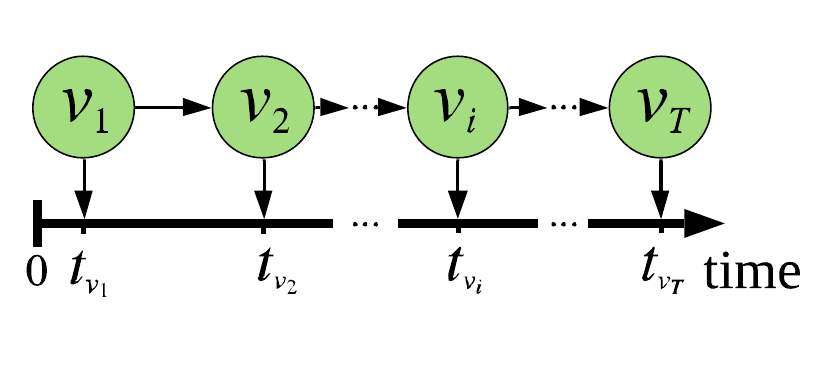}}
  \def\figc{\includegraphics[width=4.0in]{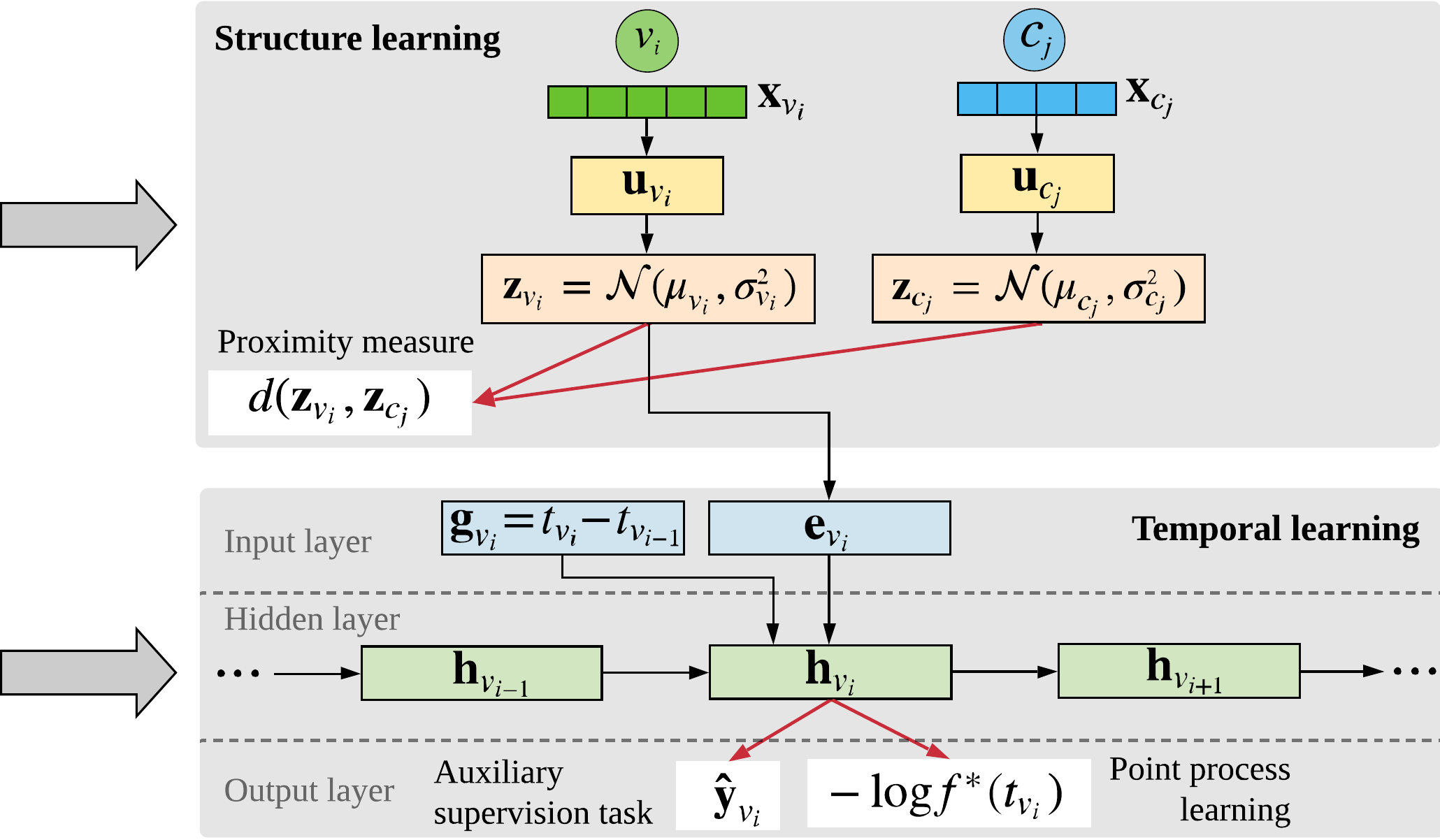}}
  \def\capa{$V$-$C$ graph
  }
  \def\capb{$V \!\!\! \xrightarrow{t} \!\! V$ sequences
  }
  \def\capc{Structural and temporal learning of $\mathtt{MedGraph}$}
  \savestack{\capfiga}{\subcaptionbox{\capa\label{fig:a}}{\figa}}
  \savestack{\capfigb}{\subcaptionbox{\capb\label{fig:b}}{\figb}}
  \savestack{\capfigc}{\subcaptionbox{\capc\label{fig:c}}{\figc}}
  \def\hgap{2ex}
  \stackon%
    [\heightof{\figc}-\heightof{\figb}-\heightof{\capfiga}-\depthof{\capfiga}]%
    {\capfigb}{\capfiga}\hspace{\hgap}\capfigc%
  \caption{$\mathtt{MedGraph}$ architecture\label{fig:archi}}
\end{figure*}

As denoted in Fig.~\ref{fig:archi}, assume a patient has a time ordered sequence of visits $v_1, v_2, \dots, v_{T}$, where $T \geq 1$ is the length of the visit sequence and each visit $v_i$ has an unordered set of medical codes $C_{v_i} \subseteq C$ with the code set $C = \{c_1, c_2, \dots, c_{|C|}\}$. $C$ denotes the set of code nodes, such as diagnosis codes.
We construct the graph data structure for this data as follows.

\textbf{$V$-$C$ attributed bipartite graph:} 
This graph has two node partitions, visits $V = \{v_1, v_2, \dots v_T\}$ and codes $C$, and a set of edges, $E_{vc}$, where $(v_i, c_j) \in E_{vc}$ is an edge denoting the link between visit $v_i$ and code $c_j$.
Each visit $v_i$ has a $D_v$-dimensional visit attribute vector $\mathbf{x}_{v_i} \in \mathbb{R}^{D_v}$ with visit demographic and utilisation information, such as age, gender and length of stay.
Each code $c_j$ has a $D_c$-dimensional code attribute vector $\mathbf{x}_{c_j} \in \mathbb{R}^{D_c}$ with code supplementary medical information, such as ICD code description as text or multi-hot ICD ontology ancestors.

\textbf{$V \xrightarrow{t} V$ temporal sequence:} 
The temporally ordered sequence of visits of a patient is denoted as $S = \{(v_i, t_{v_i}, \mathbf{y}_{v_i})\}_{i=1}^T$, where for visit $v_i$, $t_{v_i} \geq 0$ is the timestamp and $\mathbf{y}_{v_i} \in \{0,1\}^s$ is the ground-truth for the underlying auxiliary task (optional) with $s$ classes.

Each visit and code, $k \in V \cup C$, is represented as a low-dimensional Gaussian embedding in a shared embedding space, $\mathbf{z}_{k} = \mathcal{N}({\mu}_{k}, {\sigma}^2_{k})$ where ${\mu}_{k} \in \mathbb{R}^L$, ${\sigma}^2_{k} \in \mathbb{R}^{L \times L}$ with embedding dimension $L \ll |V|, |C|, D_v, D_c$ capturing visit-code associations and time-gap-based influence of historical visits.
We learn ${\sigma}_k$ as a diagonal covariance vector, ${\sigma}_k^2 \in \mathbb{R}^L$, instead of a covariance matrix to reduce the number of parameters to learn.

\subsection{$\mathtt{MedGraph}$ Architecture}

Figure \ref{fig:archi} shows the architecture of our proposed algorithm, $\mathtt{MedGraph}$.
Given an EMR dataset, we transform these data into a graph-based representation.
Then, we consider the $V$-$C$ attributed bipartite graph and learn visit and code similarities based on the graph structure proximity.
We further improve the visit embeddings by learning temporal visit sequences, $V \xrightarrow{t} V$, as events occurring in continuous-time modelled through a temporal point process which captures the mutual excitation phenomenon among temporally ordered events~\cite{hawkes_process}, so that we can learn the varying influence (due to time gaps between visits) of the historical visits.
$\mathtt{MedGraph}$ is an end-to-end risk prediction tool with a supervised task plugged into the output layer of the RNN.
Alternatively, if there is no specific supervision task, $\mathtt{MedGraph}$ can also be learned in an unsupervised manner with only structure learning and temporal sequence learning.

\subsubsection{Structural learning for $V$-$C$ graph (Fig.~\ref{fig:a})} 
In contrast to previous EMR embedding methods which collect codes in a visit as a multi-hot vector, $\mathtt{MedGraph}$ employs a bipartite graph to denote edges between $V$ and $C$. 
Due to this versatile structure, we can plug in auxiliary attribute data at nodes resulting in an attributed bipartite graph.

Let $(v_i, c_j) \in E_{vc}$ be an edge between visit $v_i$ and code $c_j$ (i.e.\ $c_j \in C_{v_i}$) with attributes $\mathbf{x}_{v_i} \in \mathbb{R}^{D_v}$ and $\mathbf{x}_{c_j} \in \mathbb{R}^{D_c}$, respectively.
These attribute vectors can be numerical values and/or one-hot encoded categorical values.
We project the two node types to a uniform semantic latent space using two transformation matrices denoted by, $\mathbf{W}_v \in \mathbb{R}^{D_v \times m}$ and $\mathbf{W}_c \in \mathbb{R}^{D_c \times m}$, where $m$ is the intermediate vector dimension, for visit and code domains, respectively.
\begin{equation}
    \mathbf{u}_{v_i} = \mathbf{W}_v \mathbf{x}_{v_i} \; \text{ and } \; \mathbf{u}_{c_j} = \mathbf{W}_c \mathbf{x}_{c_j} 
    \label{eq:u_transform}
\end{equation}
Then, we apply another layer of linear transformations to the $m$-dimensional vectors to obtain Gaussian embeddings in a common embedding space for the two node types denoted by $\mathbf{z}_{v_i} = \mathcal{N}({\mu}_{v_i}, {\sigma}^2_{v_i})$ and $\mathbf{z}_{c_j} = \mathcal{N}({\mu}_{c_j}, {\sigma}^2_{c_j})$:
\begin{align}
    {\mu}_{v_i} &= \mathbf{W}_\mu\mathbf{u}_{v_i} + \mathbf{b}_\mu \label{eq:v_mu}\\
    {\sigma}_{v_i}^2 &= ELU(\mathbf{W}_{\scriptsize \sigma}\mathbf{u}_{v_i} + \mathbf{b}_{\scriptsize \sigma}) + 1 \label{eq:v_sigma}\\
    {\mu}_{c_j} &= \mathbf{W}_\mu\mathbf{u}_{c_j} + \mathbf{b}_\mu \label{eq:c_mu}\\
    {\sigma}_{c_j}^2 &= ELU(\mathbf{W}_{\scriptsize \sigma}\mathbf{u}_{c_j} + \mathbf{b}_{\scriptsize \sigma}) + 1 \label{eq:c_sigma}
\end{align}
where $\mathbf{W}_\mu \in \mathbb{R}^{m \times L}$, $\mathbf{b}_\mu \in \mathbb{R}^{L}$, $\mathbf{W}_\sigma \in \mathbb{R}^{m \times L}$ and $\mathbf{b}_\sigma \in \mathbb{R}^{L}$ denote the shared mean and variance encoders for both node types. 
We adopted exponential linear unit (ELU) for the activation function in $\sigma$ as it drives the mean of the activation outputs be closer to zero which makes learning and convergence much faster. 
To obtain positive covariance for interpretability of uncertainty, we add one in $\sigma^2$ functions.

Embedding into a common $L$-dimensional embedding space enables similarity computation between two heterogeneous nodes.
Since the embeddings are Gaussians, we measure the Wasserstein distance as in DVNE~\cite{DBLP:conf/kdd/dvne} and RASE~\cite{DBLP:conf/pakdd/RASE}, specifically $2$-nd Wasserstein distance ($W_2$) between the embeddings.
By computing $W_2$ distance, we can preserve transitivity property in the embedding space~\cite{DBLP:conf/kdd/dvne}.
As a result, when we model the explicit visit-code relations, visit-visit and code-code associations are also implicitly modelled in the embedding space.
For example, if both $v_1$ and $v_2$ are linked to $c_2$, then it is highly likely that $v_1$ and $v_2$ are similar, and we implicitly capture this similarity by preserving triangle inequality property in the embedding space.
We define $d(\mathbf{z}_{v_i}, \mathbf{z}_{c_j})$ as the $W_2$ distance for our embeddings of visit $v_i$ and code $c_j$ in the embedding space.
Modelling only the diagonal covariance vectors results in ${\sigma}_{v_i}^2{\sigma}_{c_j}^2 = {\sigma}_{c_j}^2{\sigma}_{v_i}^2$~\cite{DBLP:conf/kdd/dvne}.
Therefore, the distance computation~\cite{givens1984classW2} simplifies to:
\begin{equation}
    \label{eq:w2}
    d(\mathbf{z}_{v_i}, \mathbf{z}_{c_j}) = W_2(\mathbf{z}_{v_i}, \mathbf{z}_{c_j}) =(\parallel {\mu}_{v_i} - {\mu}_{c_j} \parallel^2_2 + \parallel {\sigma}_{v_i} - {\sigma}_{c_j} \parallel^2_F)^{\scriptsize 1/2}
\end{equation}
Then, we define joint probability between the two node distribution representations as the likelihood of the existence of a link between them by:
$P(v_i, c_j) = Sigmoid(-d(\mathbf{z}_{v_i}, \mathbf{z}_{c_j})))$ similarly to the first-order proximity measure in GLACE~\cite{DBLP:conf/adc/GLACE}.
Since our graph is unweighted, we can define the prior probability using the structural information observed in the graph as: ${\hat{P}}(v_i, c_j) = \frac{1}{|E_{vc}|}$.
To preserve this proximity measure in the embedding space, we minimise the distance between the prior and observed probability distributions for all edges observed in $V$-$C$ graph. 
Since $\hat{P}$ and $P$ are discrete probability distributions, we define the structural loss function with Kullback–Leibler divergence ($D_{KL}$) as: 
\begin{equation}
    \label{eq:L_struct}
    \begin{split}
        \mathcal{L}_{struc} &= D_{KL}(\hat{P} || P) 
            = \!\!\!\! \sum_{(k,l) \in E_{vc}} \!\!\!\! \hat{P}(k,l) \log \Big({\frac{\hat{P}(k,l)}{P(k,l)}}\Big) \\
            &\propto -\sum_{(k,l) \in E_{vc}} \!\!\!\! \hat{P}(k,l) \log P(k,l) \\
            &\propto -\sum_{(k,l) \in E_{vc}} \!\!\!\! \log P(k,l)
    \end{split}
\end{equation}

\subsubsection{Temporal learning for $V \xrightarrow[]{t} V$ sequences (Fig.~\ref{fig:b})}

The objective of our temporal sequence modelling is to learn time gap-based influence of historical visits on the next visit of a patient.
Thus, we model the visits as continuous-time events with a point process model~\cite{hawkes_process,kingman2005poisson,point_processes} to capture the varying historical visit influence.
Typical parametric point process models establish strict assumptions on the generative process of the events, so that these models have restricted expressive power and do not guarantee to reflect the real-world data.
Hence, following the idea of RMTPP~\cite{DBLP:conf/kdd/rmtpp}, we learn the visit influence using an RNN-based architecture to model a flexible and expressive marked temporal point process and capture the visit sequence dynamics automatically without being restricted to strict parametric assumptions.

RNNs use output from the hidden units of the current time step as inputs for the next time step. 
Consequently, the network can memorise the influence of each past data event through the hidden state vectors $\mathbf{h}_{v_i}$. 
Thus, we use $\mathbf{h}_{v_i}$ to represent the influence of the history up to the $v_{i}$-th visit and model conditional intensity function of a temporal point process.
We construct a latent vector $\mathbf{e}_{v_i} \in \mathbb{R}^L$ to model the markers of the marked temporal point process using the learned visit embedding from the previous section (structural learning) $\mathbf{z}_{v_i} = \mathcal{N}({\mu}_{v_i}, {\sigma}^2_{v_i})$ where we treat the covariance vector as a noise:
\begin{equation}
    \mathbf{e}_{v_i} = {\mu}_{v_i} + \varepsilon_{v_i} \cdot {\sigma}^2_{v_i} \; \text{ where } \; \varepsilon_{v_i} \sim \mathcal{N}(\mathbf{0},\mathbf{I})
\end{equation}

For the input layer of the RNN cell (bottom block in Fig.~\ref{fig:c}), we feed the event information (i.e.\ $\mathbf{e}_{v_i}$) and the timing information (i.e.\ $\mathbf{g}_{v_i}$) about the current visit event.
Since we are interested in modelling time gaps between the visits, we set $\mathbf{g}_{v_i} = (t_{v_i} - t_{v_{i-1}})$ as the time gap between the previous and the current visit.
We update the RNN cell to output an effective hidden state vector at the current time step, $\mathbf{h}_{v_i}$, using the current visit event $(\mathbf{e}_{v_i}, \mathbf{g}_{v_i})$ and the influence from the memory carried out from the past visit events ($\mathbf{h}_{v_{i-1}}$):
\begin{equation}
    \mathbf{h}_{v_i} = ReLU(\mathbf{W}_{tv}\mathbf{e}_{v_i} + \mathbf{W}_{g}\mathbf{g}_{v_i} + \mathbf{W}_{h}\mathbf{h}_{v_{i-1}} + \mathbf{b}_h) 
    \label{eq:rnn_h}
\end{equation}
where $\mathbf{W}_{tv} \in \mathbb{R}^{L \times m^\prime}$, $\mathbf{W}_g \in \mathbb{R}^{D_t \times m^\prime}$, $\mathbf{W}_h \in \mathbb{R}^{m^\prime \times m^\prime}$, $\mathbf{b}_h \in \mathbb{R}^{m^\prime}$, $D_t$ is the time vector dimension and $m^\prime$ is the RNN hidden state dimension.
We define the conditional intensity function to model the point process by:
\begin{equation}
    \lambda^{*}(t) = \exp (\mathbf{v}_t^{\top} \cdot \mathbf{h}_{v_i} + w_t(t-t_{v_i}) + b_t) 
    \label{eq:lambda}
\end{equation}
where $\mathbf{v}_t \in \mathbb{R}^{m^\prime}$ and $w_t, b_t \in \mathbb{R}$. 
The exponential function is a non-linear transformation which guarantees positive intensity values. 
The first term $\mathbf{v}_t^{\top} \cdot \mathbf{h}_{v_i}$ comprises the \emph{historical influence}, the second term $w_t(t-t_{v_i})$ denotes the \emph{current influence}, and the final term $b_t$ emphasises \emph{base intensity} in the intensity function.
Then, we define the likelihood of the next visit occurring at time $t$ given the historical visit sequence up to time $t_{v_i}$ by:
\begin{align}
f(t - &t_{v_i} | \mathbf{h}_{v_i}) = f^{*}(t) = \lambda^{*}(t) \exp \bigg( - \int_{t_{v_i}}^{t} \lambda^{*}(\tau) d \tau \bigg)\nonumber\\
\begin{split}
    &= \exp \bigg\{  \mathbf{v}_t^{\top} \cdot \mathbf{h}_{v_i} + w_t(t-t_{v_i}) + b_t  \\
    & \;\;\;\; + \frac{1}{w_t} \bigg( \exp (\mathbf{v}_t^{\top} \cdot \mathbf{h}_{v_i} + b_t) - \exp(\mathbf{v}_t^{\top} \cdot \mathbf{h}_{v_i} + w_t(t-t_{v_i}) + b_t) \bigg) \bigg\}
\end{split}
\label{eq:f} 
\end{align}

Accordingly, given a temporal visit sequence $S$ for a patient with $T$ visits, we can define the temporal loss function as the negative log-likelihood of observing the visit sequence (i.e.\ maximise the likelihood of observing the sequence by minimising the negation):
\begin{equation}
    \mathcal{L}_{temp}   = - \sum_{i=1}^{T} \log f(t_{v_{i+1}} - t_{v_i} | \mathbf{h}_{v_i}) 
    \label{eq:L_temp}
\end{equation}


\subsubsection{Auxiliary supervision task}\label{sec:l_task}

We can train $\mathtt{MedGraph}$ to predict medical risk in the future.
Medical risk prediction is an important task for personalised healthcare~\cite{DBLP:conf/nips/retain,DBLP:conf/nips/mime}. 
Therefore, we incorporate an auxiliary medical risk prediction task.  This allows prediction of future outcomes of a patient at a given point in time to supplement predictive personalised healthcare.
We assume that the hidden state of the current visit, $\mathbf{h}_{v_i}$, not only carries the information from the current visit itself, but also memorises the time-gap-based influence of past visits through point process modelling.
For simplicity, we describe a classification task in which the outcome for the visit is $\mathbf{y}_{v_i} \in \{0, 1\}^{s}$ where $s$ is the number of classes.
We use $\mathbf{h}_{v_i}$ to predict the class label (i.e.\ future medical risk outcome) as follows:
\begin{equation}
    \mathbf{\hat{y}}_{v_i} = Softmax(\mathbf{W}_s \mathbf{h}_{v_i} + \mathbf{b}_s) 
    \label{eq:sup}
\end{equation}
where $\mathbf{W}_s \in \mathbb{R}^{s \times m^\prime}$ and $\mathbf{b}_s \in \mathbb{R}^s$.
Then, we compute the classification loss for a visit sequence $S$ of a patient using cross-entropy:
\begin{equation}\small
    \mathcal{L}_{tsk} = - \frac{1}{T} \sum_{i=1}^{T} \bigg( \mathbf{y}_{v_i}^\top \log (\mathbf{\hat{y}}_{v_i}) + (1 - \mathbf{y}_{v_i})^\top \log (\mathbf{1 - \mathbf{\hat{y}}}_{v_i}) \bigg)  
    \label{eq:L_sup}
\end{equation}

\subsection{Unified Training and Model Optimisation}

$\mathtt{MedGraph}$ is an end-to-end medical risk prediction model, which exploits $V$-$C$ graph structure and $V \xrightarrow{t} V$ temporal sequences in improving predictive performance of an underlying medical risk prediction task.
For each patient, the unified loss of the predictive model is defined as:
\begin{equation}
    \mathcal{L} = \alpha \mathcal{L}_{struc} + \beta \mathcal{L}_{temp} + \gamma \mathcal{L}_{tsk}
    \label{eq:L_sup_all}
\end{equation}
where $\alpha, \beta, \gamma \geq 0$ are hyperparameters which control learning from structural, temporal and underlying healthcare prediction task, respectively.
$\mathtt{MedGraph}$ can be trained in an unsupervised manner (setting $\gamma = 0$) when there is no auxiliary supervision task, e.g.\ when learning general-purpose embeddings for an exploratory analysis of EMR data~\cite{DBLP:conf/kdd/med2vec}.


To optimise the structural loss computation, we use the negative sampling~\cite{DBLP:conf/nips/word2vec,DBLP:conf/www/line} approach, which selects $K$ number of negative $V$-$C$ edges for each positive edge. 


\section{Experiments}

In this section, we evaluate the performance of $\mathtt{MedGraph}$ on two real-world EMR datasets and compare its performance against several state-of-the-art embedding methods along with two variant versions of $\mathtt{MedGraph}$ in the ablation study.
We perform several medical risk prediction tasks, qualitative analysis of the medical code representations and the uncertainty modelling capability of the embeddings produced by $\mathtt{MedGraph}$. 
Source code for $\mathtt{MedGraph}$ is available at \url{https://github.com/bhagya-hettige/MedGraph}.

\subsection{Datasets}

We use two real-world, cohort-specific proprietary EMR datasets in the experimental study: heart failure (HF) and chronic liver disease (CL).
We remove patients with less than two visits.
Brief statistics of the two datasets after removing the shorter visit sequences are shown in Table~\ref{tab:data_stat}.
Both EMR datasets are extracted from the same hospital,
in which \textit{ICD-10-CM diagnosis codes} and \textit{in-house procedure codes} are used. 
For the visits we extract patient demographics (e.g.\ age, gender, ethnicity, birth country, etc.) and hospitalisation utilisation information (e.g.\ length of stay, admission source, etc.) as visit attributes. 
For the medical codes, we use tf-idf vectors of ICD-10-CM code descriptions for the diagnoses, and tf-idf vectors of code descriptions provided by the hospital for the in-house procedures as code attributes. 

\begin{table}
    \begin{center}
    {\caption{Statistics of the real-world cohort-specific EMR datasets.}\label{tab:data_stat}}
        \begin{tabular}{lrr}
            \hline
            \rule{0pt}{12pt}
            Dataset & HF & CL \\
            \hline
            \\[-6pt]
            Data collection time period & 2010-2017 & 2000-2017 \\
            Total \# of patients & 10,713 & 3,830 \\
            Total \# of visits & 204,753 & 122,733 \\
            Avg. \# of visits per patient & 18.51 & 29.84 \\
            \hline
            Total \# of unique medical codes & 8,541 & 8,382 \\
            \quad \# of unique diagnosis codes & 6,278 & 6,010 \\
            \quad \# of unique procedure codes & 2,263 & 2,372 \\
            Avg.\ \# of medical codes per visit & 5.27 & 5.02 \\
            Max \# of medical codes per visit & 98 & 100 \\
            \hline
        \end{tabular}
    \end{center}
    \vspace{-3mm}
\end{table}

\subsection{Baselines}

We compare $\mathtt{MedGraph}$ to several state-of-the-art EMR embedding methods to evaluate the performance on several risk prediction tasks.
We choose Skip-gram based (Med2Vec~\cite{DBLP:conf/kdd/med2vec}) and RNN-based (Dipole~\cite{DBLP:conf/kdd/dipole} and RETAIN~\cite{DBLP:conf/nips/retain}) EMR embedding methods for comparison\footnote{GRAM~\cite{DBLP:conf/kdd/gram} is not chosen a baseline as the in-house procedure codes in our proprietary datasets do not form an ontology on which GRAM depends.}.
We also choose a state-of-the-art general-purpose graph embedding model, GCN~\cite{DBLP:conf/iclr/gcn}, to learn $V$-$C$ attributed bipartite graph.

\textbf{Med2Vec}~\cite{DBLP:conf/kdd/med2vec} is a Skip-gram based EMR embedding method that produces both code- and visit-level representations by predicting medical codes appearing in neighbouring visits. It captures visit demographics, and we feed the visit attribute vector $\mathbf{x}_{v_i}$ for each visit.

\textbf{Dipole}~\cite{DBLP:conf/kdd/dipole} is an attention-based bidirectional RNN framework, which takes the influence of historical visits via the trained attention weights.

\textbf{RETAIN}~\cite{DBLP:conf/nips/retain} is an end-to-end RNN-based healthcare prediction model with a reverse-time attention mechanism, which models the influence of previous visits and important medical codes in them. 

\textbf{GCN}~\cite{DBLP:conf/iclr/gcn}, the graph convolutional networks (GCN) model, is a recent state-of-the-art semi-supervised graph embedding approach which learns by aggregated neighbourhood information. 
We use GCN layers (unsupervised) to model the $V$-$C$ bipartite relations.
Since GCN only supports homogeneous graphs, we ignore the node heterogeneity and construct attribute vectors of visits (i.e.\ $\mathbf{x}_{v_i}$) and codes (i.e.\ $\mathbf{x}_{c_j}$) by pre- and post-padding with zeros respectively. 

\subsection{Our Approaches}

In addition to these baseline methods, we conduct a comprehensive ablation study to evaluate the effectiveness and importance of the two important components of $\mathtt{MedGraph}$, namely structure learning and point process based temporal learning. 
We denote each variant model with a negation ($\lnot$) in front of the ablated component in learning.
We summarise the variants in Table~\ref{tab:medgraph_approaches}.

\begin{table}
    \begin{center}
        \caption{Our approaches trained with Eq.~\ref{eq:L_sup_all}:\\
        $\mathcal{L} = \alpha \mathcal{L}_{struc} + \beta \mathcal{L}_{temp} + \gamma \mathcal{L}_{tsk}$.}
        \label{tab:medgraph_approaches}
        \begin{tabular}{lccc}
            \hline
            \rule{0pt}{12pt}
            Notation & $\alpha$ & $\beta$ & $\gamma$ \\
            \hline
            \\[-6pt]
            $\mathtt{MedGraph}$ & $>0$ & $>0$ & \multirow{3}{*}{\parbox{2.8cm}{$>0$ for the predictive model; $=0$ for the unsupervised model}} \\
            $\mathtt{MedGraph{(S,\neg T)}}$ & $>0$ & $=0$ & \\ 
            $\mathtt{MedGraph{(\neg S,T)}}$ & $=0$ & $>0$ & \\
            \hline
        \end{tabular}
    \end{center}
\end{table}

\textbf{\texttt{MedGraph}} is our full model that incorporates both structural and temporal aspects (Fig.~\ref{fig:archi}) by modelling $V$-$C$ bipartite relations and time-gap-based temporal point process of visit sequences.

\textbf{\texttt{MedGraph(S,$\neg$T)}} is a simple RNN model which treats visit events as an equally spaced time-series and thus does not model temporal point processes. It learns from $V$-$C$ bipartite associations.

\textbf{\texttt{MedGraph($\neg$S,T)}} does not learn any structural information in the $V$-$C$ graph, so it does not learn medical codes of visits. It implements the proposed temporal point process based RNN model to learn time-gap-based influence of historical visits.

\subsection{Hyperparameter Settings}
For all baseline models, we use $L=256$ as the visit and code embedding dimension.
Since $\mathtt{MedGraph}$ produces mean and covariance vectors, we halve the embedding dimension to $L=128$ in $\mathtt{MedGraph}$ for a fair comparison by learning the same number of parameters per node (i.e.\ ${\mu}_i \in \mathbb{R}^{128}$ and ${\sigma}^2_i \in \mathbb{R}^{128}$).
For all the methods, we use a batch size of 128 for visits and 32 for visit sequences. 
For $\mathtt{MedGraph}$, we set the number of negative edges as 10, and $\alpha, \beta, \gamma$ are tuned to be optimal on a validation set. 
We use Adam optimiser with a learning rate fixed at 0.001. 
Other parameters for baseline models are referred from the papers and tuned to be optimal. 
$\mathtt{MedGraph}$'s RNN layer uses LSTM.
All the experiments are conducted on a modest hardware setting with a MacBook Pro laptop with 16GB memory and a 2.6 GHz Intel Core i7 processor.

\subsection{Medical Risk Prediction}

We perform two real-world medical risk prediction tasks, 30-day readmission prediction and mortality prediction, to assess the model's effectiveness in predictive healthcare.
For $\mathtt{MedGraph}$, we use each risk prediction task as the auxiliary supervision task (cf Section~\ref{sec:l_task}) in this set of experiments in an end-to-end manner ($\gamma > 0$). 
Since Med2Vec, Dipole and GCN are not designed for medical risk prediction tasks, we first learn visit embeddings using these methods, and then use these fixed visit vectors as inputs to the prediction model (i.e.\ XGBoost classifier~\cite{DBLP:conf/kdd/xgboost}). 
For both tasks, we randomly split the visit sequences into 2 parts with a 4:1 ratio. 
The former is used to train the embedding models, while the latter is held out for evaluation. 
We randomly sample 25\% of visit sequences from the held-off dataset as the validation set, and the rest as the test set. 

\subsubsection{30-day readmission risk prediction} 
\begin{figure}[t]
    \centering
    \includegraphics[width=0.85\linewidth]{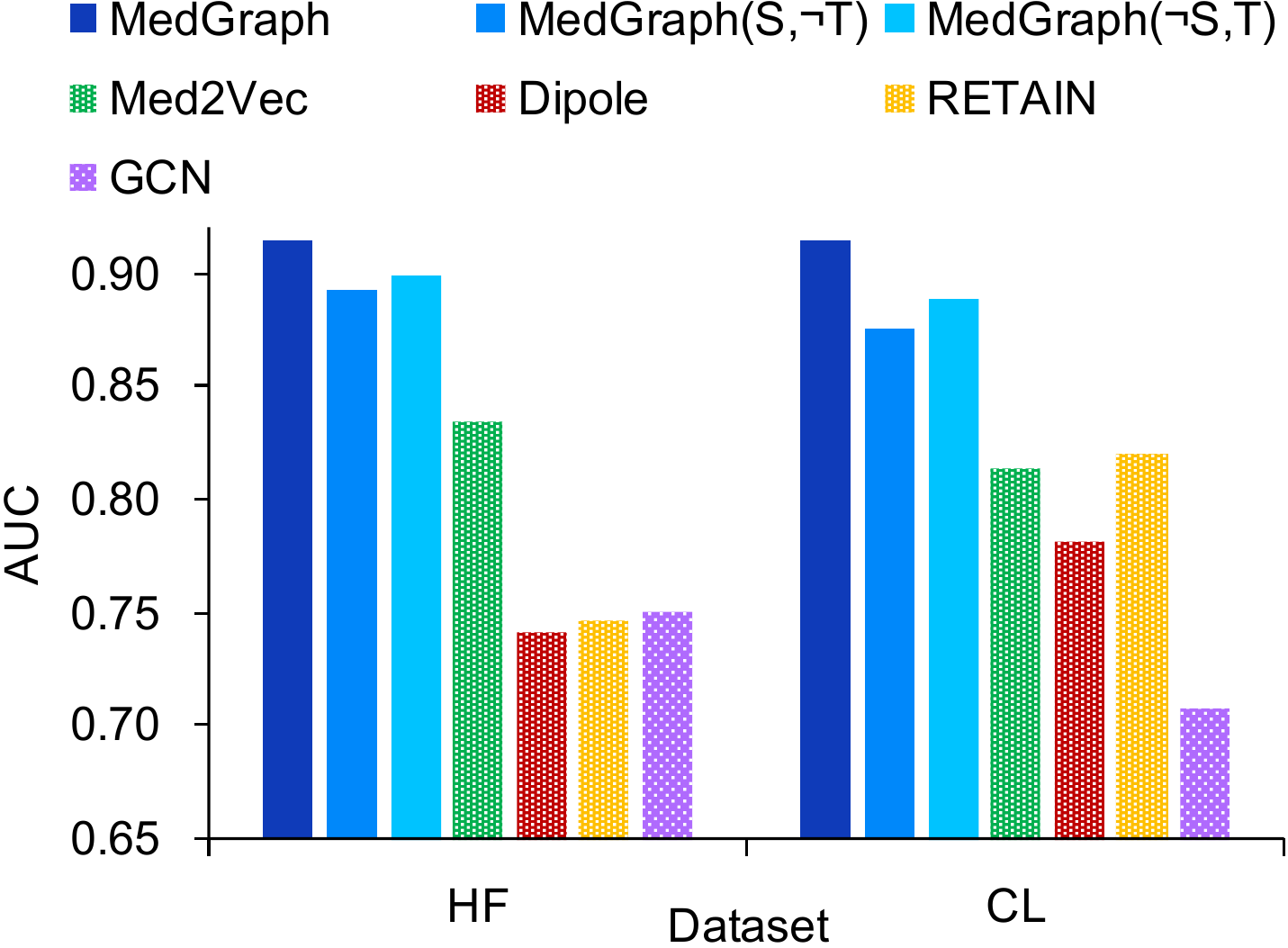}
    \caption{Performance on 30-day readmission risk prediction task. 
    }
    \label{fig:readmission}
\end{figure}

Accurate readmission prediction allows hospitals to target intervention strategies on high-risk patients to prevent readmissions. 
Given a visit, we train the models to predict 30-day readmission after a patient's discharge from the current visit/admission~\cite{journal/jama/readmission,nguyen2016mathtt_deepr}, thus casting it as a binary classification task. 
Here, we consider all the readmissions and not only the unplanned readmissions.
Both cohort datasets are imbalanced and the majority class is positive, with a prevalent 30-day readmission rate of 69.8\% and 76.2\% respectively. 
Therefore, area under the receiver operating curve (AUC) is a better measure to compare the models~\cite{DBLP:conf/nips/retain}. 
We plot AUC scores for HF and CL datasets in Figure~\ref{fig:readmission}. 

As can be seen from the plot, $\mathtt{MedGraph}$, as well as its two variants, clearly and consistently outperform all compared state-of-the-art baseline methods by a significant margin. 
This shows the effectiveness of each component in our model.
The contribution of each type of information is experimentally validated by the performance improvement gains of the full model over the non-structural ($\mathtt{MedGraph{(\neg S,T)}}$) and 
time-series
($\mathtt{MedGraph{(S,\neg T)}}$ with no time gaps) variants of our method.
Our observation is also supported by a previous study on readmissions~\cite{DBLP:conf/kdd/readmission}, which shows that readmission of a patient is dependent on various factors including patient demographics and past medical incidents.

Among the EMR baseline models, Med2Vec performs the best on HF, and second best on CL even though it does not model temporality of the visits,
which can be attributed to its
involvement of visit demographics in embedding.
Capturing visit-code associations is shown to be effective, as GCN's performance is competitive compared with several EMR embedding baselines in this task. 

From the performance gaps between RNN-based methods (Dipole, RETAIN and $\mathtt{MedGraph{(S,\neg T)}}$) and our time-gap-based point process models ($\mathtt{MedGraph}$ and $\mathtt{MedGraph{(\neg S,T)}}$), we can see that time gap information modelling is important in making an accurate prediction of the readmission risk in both datasets. 
Actually, our proposed point process model learns time gaps between historical visits via point process modelling (Eq.~\ref{eq:L_temp}).
Consequently, we can successfully predict a patient's readmission with a higher prediction accuracy at the 30-day time threshold. 

\subsubsection{Mortality prediction} 
Early identification of patients who are at high risk of death can assist hospitals to appropriately allocate resources to these patients and mitigate the mortality risk. 
Given a patient's visit sequence, we train the models to predict the patient's death in the next admission~\cite{knaus1991apache}, thus casting it as a binary classification task. 
We report area under receiver operating curve (AUC) and 
average precision (AP) on 
both HF and CL datasets (with a prevalent mortality rate of 27.0\% and 40.7\% respectively) in Table \ref{tab:mortality prediction}. 

From the table, we can see that $\mathtt{MedGraph}$ shows superior performance over all the other methods across both datasets. 
This demonstrates the effectiveness of the visit representations learned by $\mathtt{MedGraph}$ in predicting the death of patients in their next visit.
The possible reasons for the superiority of our method include: the incorporation of attribute information in visits and codes and adopting temporal point processes in modelling visit-visit sequences. 
Intuitively, according to a related study~\cite{DBLP:conf/amcis/mortality}, the outcome of a patient's death depends on their demographics (i.e.\ $\mathbf{x}_{v_i}$), current health conditions (i.e.\ $V$-$C$ structure) and historical health conditions (i.e.\ $V \xrightarrow{t} V$ sequence), all of which are captured by our model.

Moreover, $\mathtt{MedGraph}$ outperforms both its variant versions, which showcases the effectiveness of the proposed structural and temporal learning in improving predictive performance of the visit representations in the mortality prediction task.

The poor performance of GCN and the performance gain of the RNN-based methods (i.e.\ Dipole and RETAIN) over GCN, suggest that the structural learning is not sufficient to capture visit similarities in this task and temporality is important.
However, Med2Vec performs better due to Skip-gram based code co-occurrence learning and incorporation of visit attributes, though it ignores temporality.



\begin{table}
    \begin{center}
    {\caption{Performance on the mortality prediction task. Best result is \textbf{bolded} and second-best is \underline{underlined}.}\label{tab:mortality prediction}}
        \begin{tabular}{clrrrr}
            \hline
            \rule{0pt}{12pt}
            & & \multicolumn{2}{c}{HF} & \multicolumn{2}{c}{CL}\\
            \multicolumn{2}{c}{Method} & \multicolumn{1}{c}{AUC} & \multicolumn{1}{c}{AP} & \multicolumn{1}{c}{AUC} & \multicolumn{1}{c}{AP} \\
            \hline
            \\[-6pt]
            & Med2Vec & 0.7091 & 0.4752 & 0.7224 & 0.6255 \\
            Baselines & Dipole & 0.6464 & 0.4145 & 0.6411 & 0.5487 \\
            & RETAIN & 0.6908 & 0.5402 & 0.7034 & 0.6836 \\
            & GCN & 0.5581 & 0.3177 & 0.6172 & 0.4956 \\
            \hline
            & $\mathtt{MedGraph{(S,\neg T)}}$ & 0.7002 & 0.6548 & \underline{0.7385} & \underline{0.7123} \\
            Ours & $\mathtt{MedGraph{(\neg S,T)}}$ & \underline{0.7131} & \underline{0.6751} & 0.7306 & 0.7118 \\
            & $\mathtt{MedGraph}$ & \textbf{0.7205} & \textbf{0.6853} & \textbf{0.7415} & \textbf{0.7143} \\
            \hline
        \end{tabular}
    \end{center}
\end{table}

\subsection{Medical Code Representation Analysis}
In this section, we explore the medical code embeddings to evaluate their descriptiveness.
We propose a novel evaluation task to quantitatively analyse the informativeness of the code embeddings and we qualitatively study the interpretability of the learned code embeddings by $\mathtt{MedGraph}$.
We learn general-purpose embeddings with $\mathtt{MedGraph}$ with $\gamma = 0$ in Eq.~\ref{eq:L_sup_all}.

\subsubsection{Multi-class code classification}

The Clinical Classifications Software (CCS)\footnote{\url{https://www.hcup-us.ahrq.gov/toolssoftware/ccsr/ccs_refined.jsp}} divides the ICD codes into a number of clinically meaningful categories.
Thus, if the code embeddings are predictive of their CCS categories, then the embeddings have learnt useful latent information.
With this hypothesis, we perform multi-class code classification to predict the corresponding medical concept classes of medical codes produced by CCS. 

First, each method learns code embeddings, and then a logistic regression (LR) classifier is trained on the code embeddings to classify each code into their associated CCS class.
We select the 10 most common CCS classes in the datasets, since some CCSs are rare.
We randomly sample different percentages of diagnosis codes $(10\%, 20\%, \dots 90\%)$ as the training set for the classifier, and use the rest for evaluation. 
We report micro- and macro-F1 scores which have been widely used in the evaluation of multi-class classification tasks~\cite{DBLP:conf/www/line, DBLP:conf/kdd/dvne}. 

\begin{figure}[!t]
    \centering
    \begin{minipage}{\linewidth}
        \centering
        \includegraphics[width=0.8\linewidth]{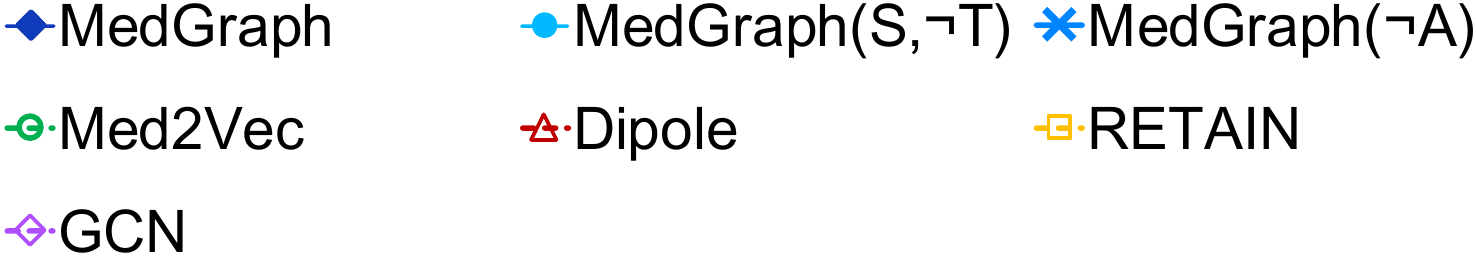}
    \end{minipage}\hfill
    \vspace{3mm}
    \begin{minipage}{\linewidth}
        \centering
        \includegraphics[width=0.85\linewidth]{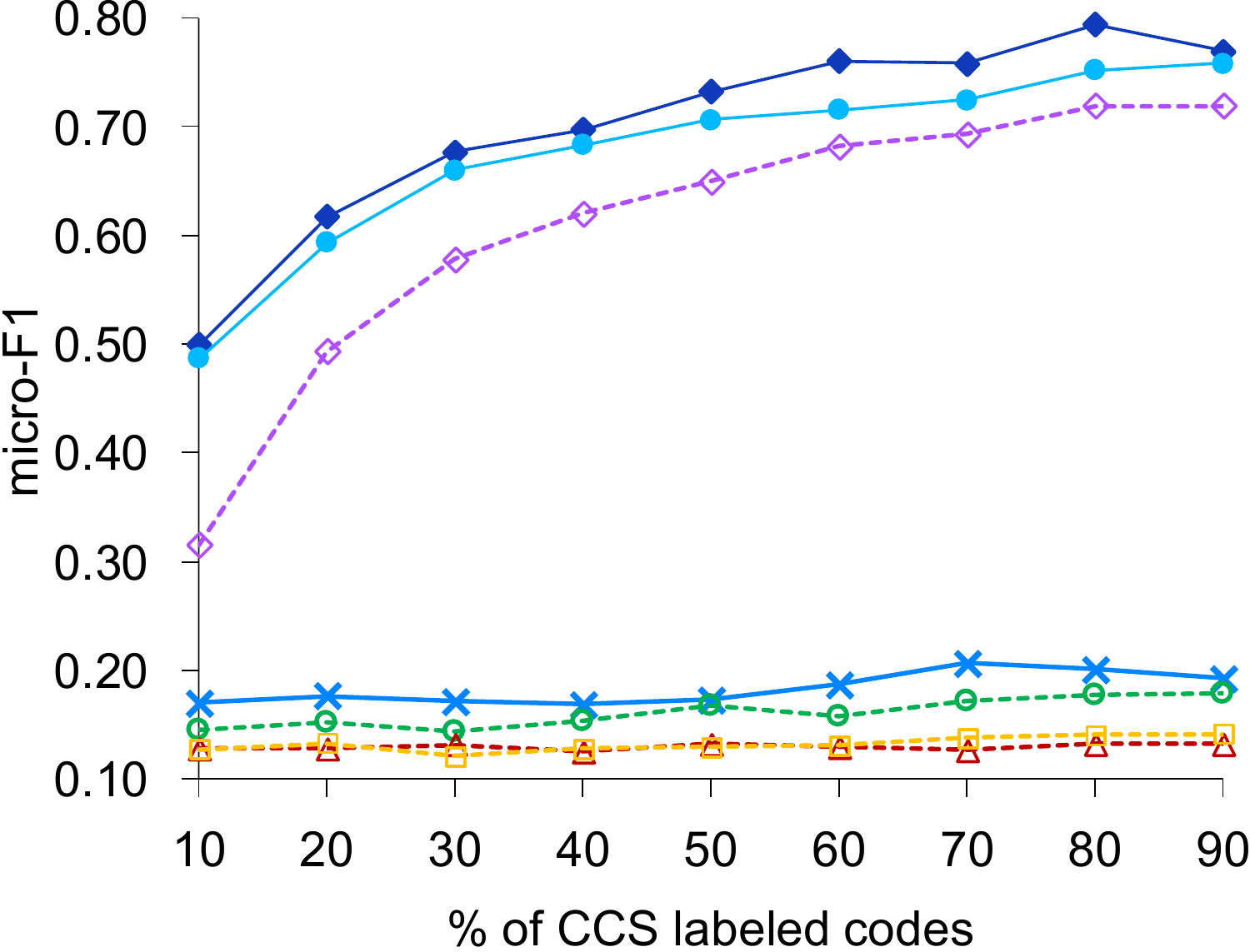}
    \end{minipage}
    \caption{Multi-class code classification task. Improvements are statistically significant for $p<0.01$ in a paired t-test.} \label{fig:code_class}
\end{figure}

$\mathtt{MedGraph}$ is capable of learning supplementary attributes of codes.
Thus, it should be able to produce more expressive code embeddings compared to the methods which do not use attributes. 
To conduct a fair evaluation with these baselines, 
we train $\mathtt{MedGraph}$ without code attributes (i.e.\ let $\mathbf{X}_C = \mathbf{I}_{|C|}$). 
This variant is denoted $\mathtt{MedGraph(\neg A)}$. 
We exclude $\mathtt{MedGraph{(\neg S,T)}}$ in this task as it does not learn code embeddings.
Micro-F1 scores for HF are presented in Figure~\ref{fig:code_class}.
The trend is similar for macro-F1 scores, and in the CL dataset, which we omit for brevity reasons. 

As can be seen in Fig.~\ref{fig:code_class}, $\mathtt{MedGraph}$ and $\mathtt{MedGraph{(S,\neg T)}}$ produce embeddings that are highly descriptive of their medical context, consistently and substantially outperforming the non-attributed version, $\mathtt{MedGraph(\neg A)}$, by a large margin of improvements in the code classification task.
GCN also showcases a significant performance improvement over non-attributed models.
This demonstrates the effectiveness of incorporating code-level attributes and structural visit-code relations in producing high quality code embeddings. 
Thus, superiority of our method over $\mathtt{MedGraph(\neg A)}$ and the rest of the baselines can be attributed to two factors: (1) the use of standard code descriptions as supplementary code attributes, and (2) the learning of code-visit associations through the graph-based data structure, as opposed to multi-hot medical code vectors.

We also see that $\mathtt{MedGraph(\neg A)}$ consistently outperforms all the three evaluated EMR embedding methods, producing more meaningful embeddings.
The difference between our $\mathtt{MedGraph(\neg A)}$ method and these methods is the way we learn visit-code associations.
These baseline methods construct multi-hot medical code vectors to represent visits, and then these are used as features in their models to learn visit-code relations ignoring the inherent graph structure.
On the contrary, our models learn visit-code associations through the structural information in the $V$-$C$ bipartite graph.
Thus, ours is capable of learning not only the code co-location in visits through local neighbourhood, but also the similar code neighbourhoods through global connectivity
due to transitivity property in $W_2$ distance.
This shows that the structural learning of visit-code associations we proposed is effective in producing meaningful code embeddings.
Among the baselines, Med2Vec shows superiority over Dipole and RETAIN, which is attributed to its neighbouring code learning technique (similar to Skip-gram) within a predefined context window.

$\mathtt{MedGraph{(S,\neg T)}}$, which learns $V$-$C$ structure and model time-series visits with no time gaps, is slightly surpassed by $\mathtt{MedGraph}$, since injecting time-gap-based temporal learning of medical history enables to learn additional useful latent patterns in EMRs. 
For example, when a patient is diagnosed with cancer at an earlier visit, $\mathtt{MedGraph}$ learns that the patient may revisit the hospital for chemotherapy or other related procedures via the patient's medical history. 

\subsubsection{Interpretation of code embeddings}

\begin{figure}[!t]
    \centering
    \includegraphics[width=\linewidth]{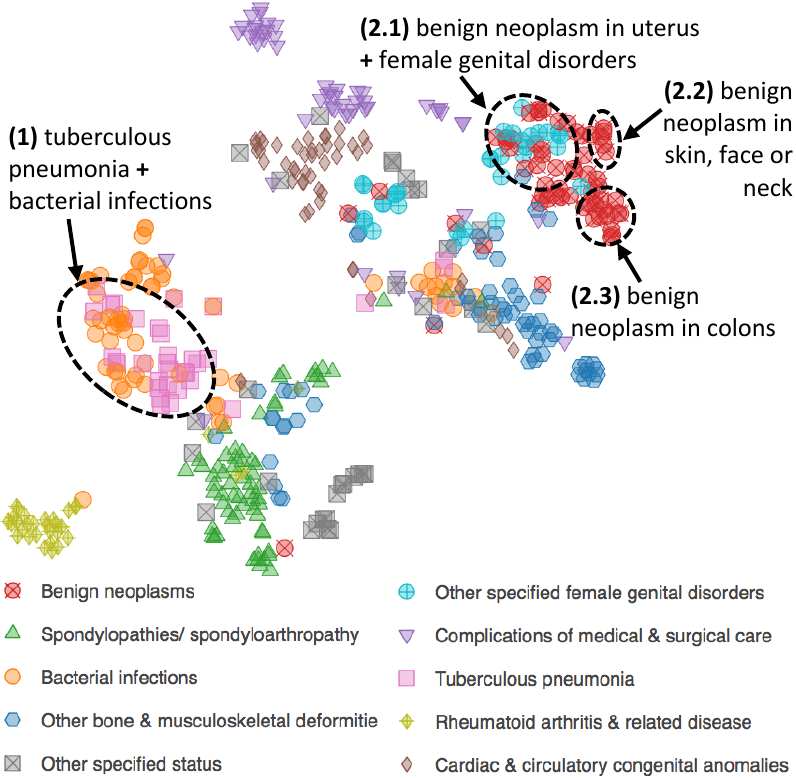}
    \caption{2-D visualisation of codes. Code's colour is its CCS class.} \label{fig:code_int}
\end{figure}

Interpretability of the learned code embeddings is important in various applications in medical domain, including healthcare analysis tasks.
In Figure~\ref{fig:code_int}, we project the 128-dimensional mean vector of embeddings of ICD-10-CM diagnosis codes trained with $\mathtt{MedGraph}$ on HF dataset ($\sim$500 codes belonging to the top 10 CCS classes) into 2 dimensions using t-SNE, for visualisation~\cite{DBLP:tsne}.
We also publish an interactive plot for further analysis\footnote{\url{https://bhagya-hettige.github.io/MedGraph}}.
Colour of a node indicates the associated CCS class. 

Overall, $\mathtt{MedGraph}$ clusters codes belonging to most CCS clinical concepts with clear boundaries.
Moreover, there are several overlapped CCS classes due to broad definitions of CCS.
$\mathtt{MedGraph}$ learns interesting latent relations between codes, especially in these overlapped CCS regions.
For brevity reasons, we only analyse two such scenarios:
(1) codes of ``tuberculosis pneumonia'' and ``bacterial infections'' are overlapping, forming a cluster showing their clinically closer relationships~\cite{10.1093/qjmed/pneumonia}, and
(2) ``benign neoplasm'' related codes further separates into more granular classes (focussed on different organs) within a broader CCS class, identifying the inherent differences in these sub classes.

\subsection{Uncertainty Modelling of EMR}

Different from the existing EMR embedding methods, $\mathtt{MedGraph}$ learns the uncertainty of visit and code embeddings as Gaussians.
In this task, we study the nature of learned uncertainty terms and its intuition in the real-world EMRs. 
We learn interpretable diagonal covariances with non-negative values (cf Section~\ref{sec:method}).
We define the average variance across the 10 largest dimensions as a node's variance~\cite{DBLP:conf/kdd/dvne}. 
We conduct the following analysis on the HF dataset.
To obtain general-purpose embeddings in this exploratory analysis, we learn the embeddings in an unsupervised manner, setting $\gamma = 0$ in Eq.~\ref{eq:L_sup_all}.

\begin{figure}[!t]
    \centering
    \begin{subfigure}{0.495\linewidth}
        \centering
        \includegraphics[width=\linewidth]{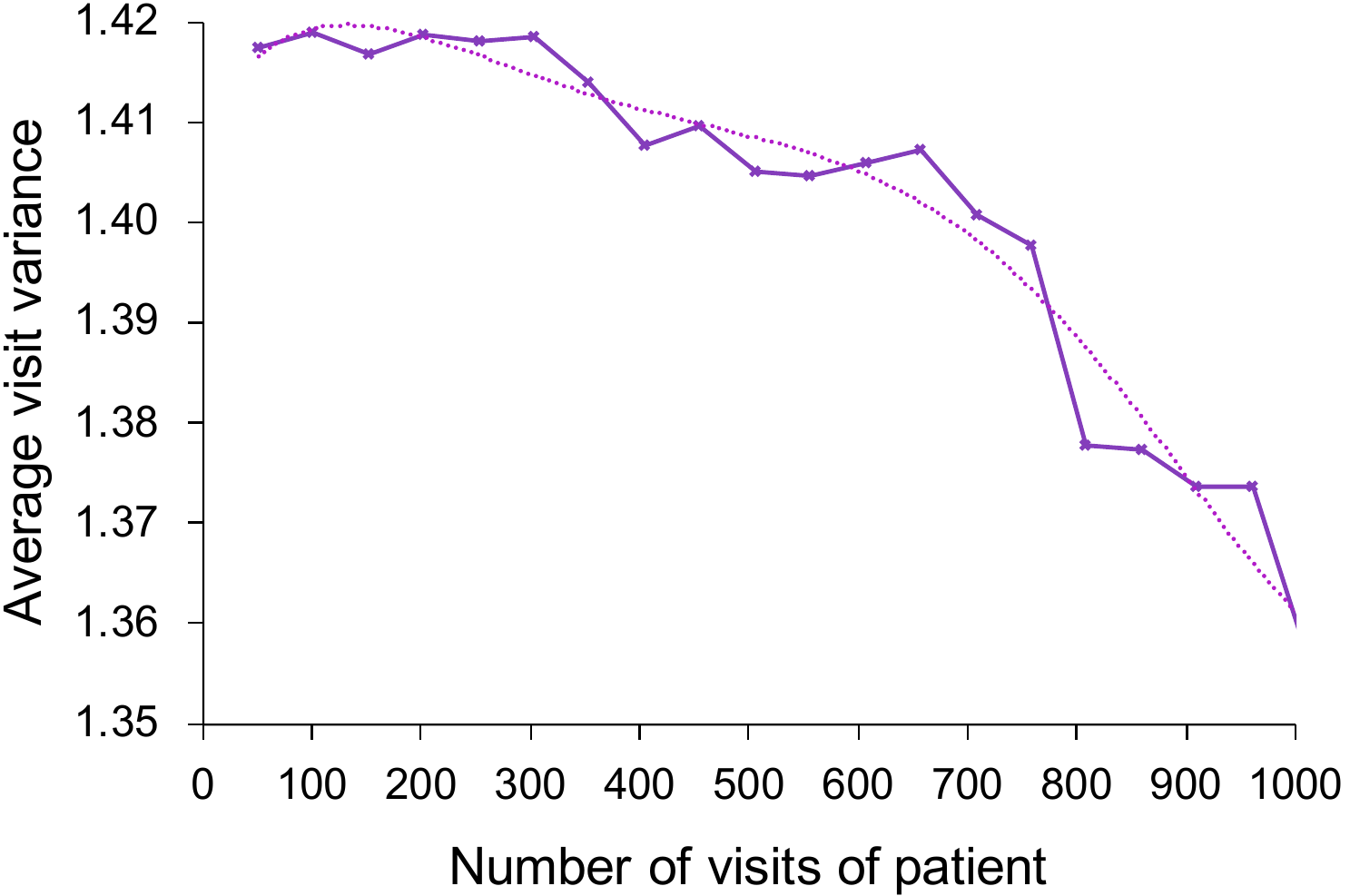}
        \caption{Visit uncertainty.}
        \label{fig:un_visits}
    \end{subfigure}
    \begin{subfigure}{0.495\linewidth}
        \centering
        \includegraphics[width=\linewidth]{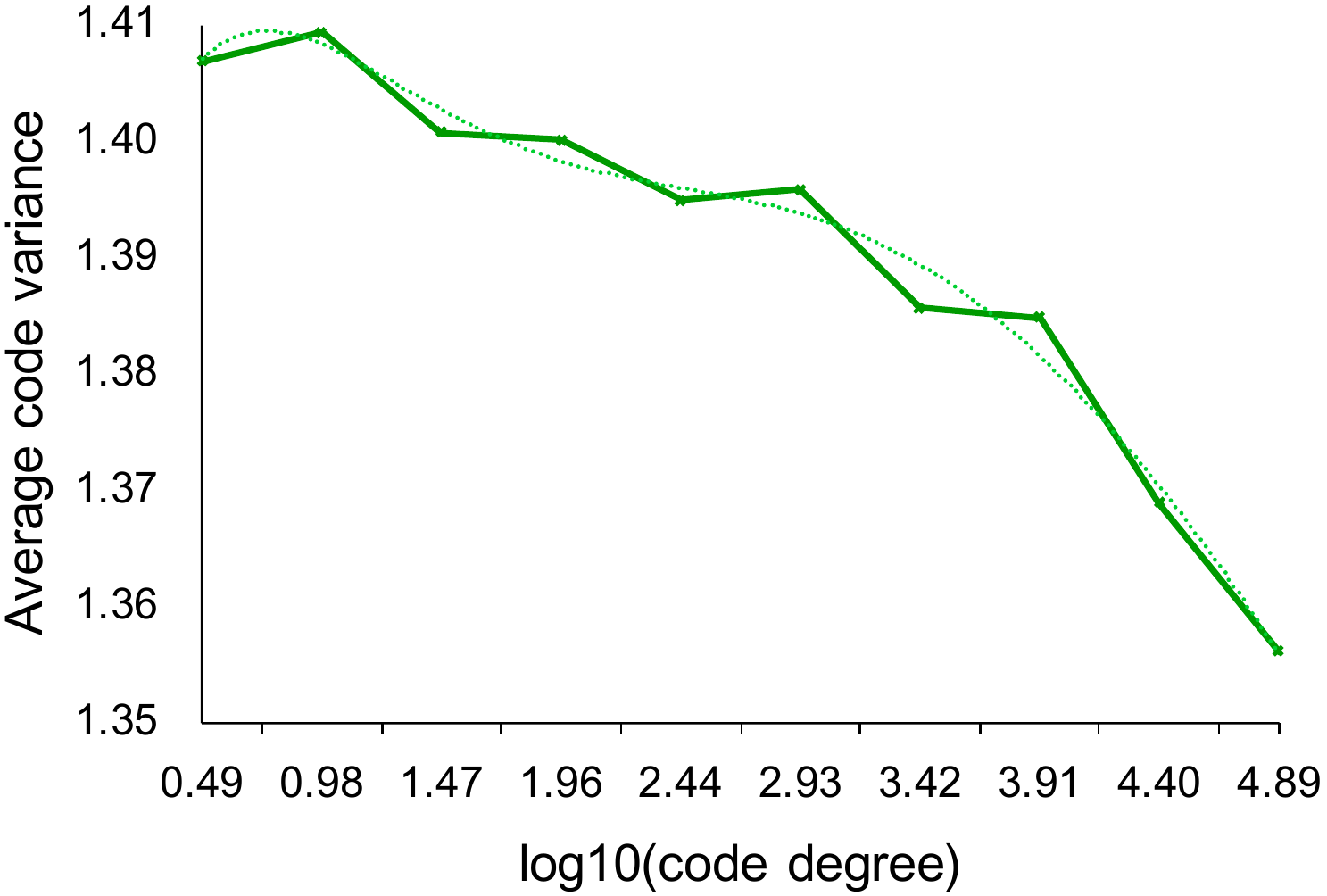}
        \caption{Code uncertainty.}
        \label{fig:un_codes}
    \end{subfigure}
    \caption{Analysis of uncertainty of the embeddings. Trendlines show the trend of the results.}
    \label{fig:uncertainty}
\end{figure}

\textbf{Visit Embedding Uncertainty (Fig.~\ref{fig:un_visits}): } 
We divide all the patients into 20 buckets based on their visit counts. 
For each bucket, we compute the average visit variance and plot it against the number of visits.
When the number of visits of a patient increases the average variance of visit embeddings decreases (Fig.~\ref{fig:un_visits}). 
Intuitively, when a patient has a longer medical history, their visit embeddings are more comprehensive and descriptive, thus less uncertain. 

\textbf{Code Embedding Uncertainty (Fig.~\ref{fig:un_codes}): } 
We divide the ICD-10
codes into 10 buckets based on their degrees (i.e.\ number of visits a code is connected to). 
We compute the average variance of each bucket and plot it against the $\log_{10}(degree(code))$.
We observe that the average variance decreases, when the code degree increases in Fig.~\ref{fig:un_codes}. 
Intuitively, lower degree codes (i.e.\ when the code rarely occurs) have less structural information to learn, hence their embeddings have a higher degree of uncertainty. 
In contrast, higher degree codes occur more frequently, so they possess a lower embedding uncertainty as they are more expressive in terms of the structure. 

\section{Conclusion}

Learning low-dimensional representations for EMRs is essential in improving personalised healthcare.
In this work, we propose $\mathtt{MedGraph}$, an effective EMR embedding framework for visits and codes. 
$\mathtt{MedGraph}$ introduces a graph-based data structure to naturally capture both visit-code co-location information \emph{structurally}, and visit sequencing information \emph{temporally}. 
Based on this structure $\mathtt{MedGraph}$ learns from the visit-code bipartite graph and exploits temporal point processes to capture medical history in an end-to-end manner.
$\mathtt{MedGraph}$ supports visit- and code-level attributes.
We further improve the expressive power of $\mathtt{MedGraph}$ by modelling uncertainty of the embeddings. 
Results on two real-world EMR datasets demonstrate that $\mathtt{MedGraph}$ produces meaningful representations for EMRs, significantly outperforming state-of-the-art EMR embedding methods on a number of medical risk prediction tasks.

\section*{Acknowledgements}
This work has been supported by the Monash Institute of Medical Engineering (MIME), Australia.

\bibliography{bibliography}

\end{document}